\documentclass[runningheads]{llncs}
\usepackage{graphicx}
\usepackage{tikz}
\usepackage{comment}
\usepackage{amsmath,amssymb} % define this before the line numbering.
\usepackage{color}
\usepackage{hyperref}
\hypersetup{hidelinks,
	colorlinks=true,
	allcolors=black,
	pdfstartview=Fit,
	breaklinks=true}
\usepackage[accsupp]{axessibility}  % Improves PDF readability for those with disabilities.

\usepackage{graphicx}
\usepackage{amsmath}
\usepackage{amssymb}
\usepackage{float}
\usepackage{subfigure}

\usepackage{multirow}
\usepackage{array}
\usepackage{booktabs}
\usepackage{graphicx}
\usepackage{verbatimbox}
\usepackage{bm}
\usepackage{caption}
\usepackage{wrapfig}
\usepackage{setspace}

\usepackage[perpage]{footmisc}
\usepackage{etoolbox}

\DefineFNsymbols*{asterisks}{*{\dagger}{***}}
\begin{document}
% \renewcommand\thelinenumber{\color[rgb]{0.2,0.5,0.8}\normalfont\sffamily\scriptsize\arabic{linenumber}\color[rgb]{0,0,0}}
% \renewcommand\makeLineNumber {\hss\thelinenumber\ \hspace{6mm} \rlap{\hskip\textwidth\ \hspace{6.5mm}\thelinenumber}}
% \linenumbers
\pagestyle{headings}
%\mainmattert
\def\ECCVSubNumber{1197}  % Insert your submission number here

\title{\emph{mc}-BEiT: Multi-choice Discretization \\for Image BERT Pre-training}
% INITIAL SUBMISSION 
%\begin{comment}
% \titlerunning{ECCV-22 submission ID \ECCVSubNumber} 
% \authorrunning{ECCV-22 submission ID \ECCVSubNumber} 
% \author{Anonymous ECCV submission}
% \institute{Paper ID \ECCVSubNumber}
%\end{comment}
%******************

% CAMERA READY SUBMISSION
%\begin{comment}
\titlerunning{\emph{mc}-BEiT}
% If the paper title is too long for the running head, you can set
% an abbreviated paper title here
%
%\thanks{Work done during internship in ARC Lab.}
\author{{Xiaotong Li{$^{1}$},Yixiao Ge{$^{2}$},Kun Yi{$^{2}$},Zixuan Hu{$^{1}$},Ying Shan{$^{2}$},Ling-Yu Duan{$^{1,3,*}$}
} \\
\institute{ {$^{1}$}Peking University, Beijing, China ~~ {$^{2}$}ARC Lab, Tencent PCG \\ {$^{3}$}Peng Cheng Laboratory, Shenzhen, China \\
\email{{\tt\small lixiaotong@stu.pku.edu.cn, \{hzxuan, lingyu\}@pku.edu.cn,\\\tt\small \{yixiaoge, yingsshan, kunyi\}@tencent.com}}\\
$^*$Corresponding Author\\
}}
% \author{First Author\inst{1}\orcidID{0000-1111-2222-3333} \and
% Second Author\inst{2,3}\orcidID{1111-2222-3333-4444} \and
% Third Author\inst{3}\orcidID{2222--3333-4444-5555}}
%
\authorrunning{X. Li et al.}
% First names are abbreviated in the running head.
% If there are more than two authors, 'et al.' is used.
%
% \institute{Princeton University, Princeton NJ 08544, USA \and
% Springer Heidelberg, Tiergartenstr. 17, 69121 Heidelberg, Germany
% \email{lncs@springer.com}\\
% \url{http://www.springer.com/gp/computer-science/lncs} \and
% ABC Institute, Rupert-Karls-University Heidelberg, Heidelberg, Germany\\
% \email{\{abc,lncs\}@uni-heidelberg.de}}
%\end{comment}
%******************
\maketitle

\begin{abstract}
Image BERT pre-training with masked image modeling (MIM) becomes a popular practice to cope with self-supervised representation learning.
A seminal work, BEiT, casts MIM as a classification task with a visual vocabulary, tokenizing the continuous visual signals into discrete vision tokens using a pre-learned dVAE. 
Despite a feasible solution, the improper discretization hinders further improvements of image pre-training.
Since image discretization has no ground-truth answers, we believe that the masked patch should not be assigned with a unique token id even if a better ``tokenizer'' can be obtained.
In this work, we introduce an improved BERT-style image pre-training method, namely \emph{mc}-BEiT, which performs MIM proxy tasks towards eased and refined multi-choice training objectives.
Specifically, the multi-choice supervision for the masked image patches is formed by the soft probability vectors of the discrete token ids, which are predicted by the off-the-shelf image ``tokenizer'' and further refined by high-level inter-patch perceptions resorting to the observation that similar patches should share their choices. 
Extensive experiments on classification, segmentation, and detection tasks demonstrate the superiority of our method, \textit{e.g.}, the pre-trained ViT-B achieves 84.1\% top-1 fine-tuning accuracy on ImageNet-1K classification, 49.2\% $\text{AP}^{b}$ and 44.0\% $\text{AP}^{m}$ of object detection and instance segmentation on COCO, 50.8\% mIOU on ADE20K semantic segmentation, outperforming the competitive counterparts. The code will be available at \href{https://github.com/lixiaotong97/mc-BEiT}{https://github.com/lixiaotong97/mc-BEiT}. 

\keywords{Self-supervised Learning, Vision Transformers, Image BERT Pre-training}
\end{abstract}

\section{Introduction}
%Introduction to recent progress of self-supervised learning.
% \footnote{Corresponding Author.}
Self-supervised pre-training \cite{beit,simclr,mocov3,ibot,vicreg,barlow,clip} is attracting emerging attention due to its effectiveness and flexibility in exploiting large-scale uncurated data, which demonstrates its superiority to supervised pre-training in a wide range of downstream applications, such as classification, detection, and segmentation, \textit{etc.} 
%The main idea in SSL is to design various pretext tasks in a self-supervised manner for model learning.  
Recently, the introduction of vision Transformers \cite{vit,deit,swinT} brings about a new revolution to self-supervised learning \cite{beit,mocov3,dino}. 

Inspired by the great success of BERT \cite{BERT} in natural language processing (NLP) tasks, masked image modeling (MIM) has been introduced for visual pre-training as a new pretext task.
It is not trivial, because one key barrier lies in that the visual signal is continuous and cannot be properly classified as is done in masked language modeling (MLM) of BERT.
A pioneer work, BEiT \cite{beit}, tackles the challenge by ``tokenizing'' continuous visual signals into discrete vision tokens resorting to a pre-learned codebook \cite{dalle},
% according to an off-the-shelf codebook, 
which plays the role of a pre-defined vocabulary in MLM. 
% The tokenizer together with the codebook is formed as a discrete VAE which is pre-trained by an autoencoding-style reconstruction task.
The pre-training objective is to predict the vision token id of the masked image patch based on its context and semantics.

\begin{figure}[t]
\begin{center}
\includegraphics[width=0.9\textwidth]{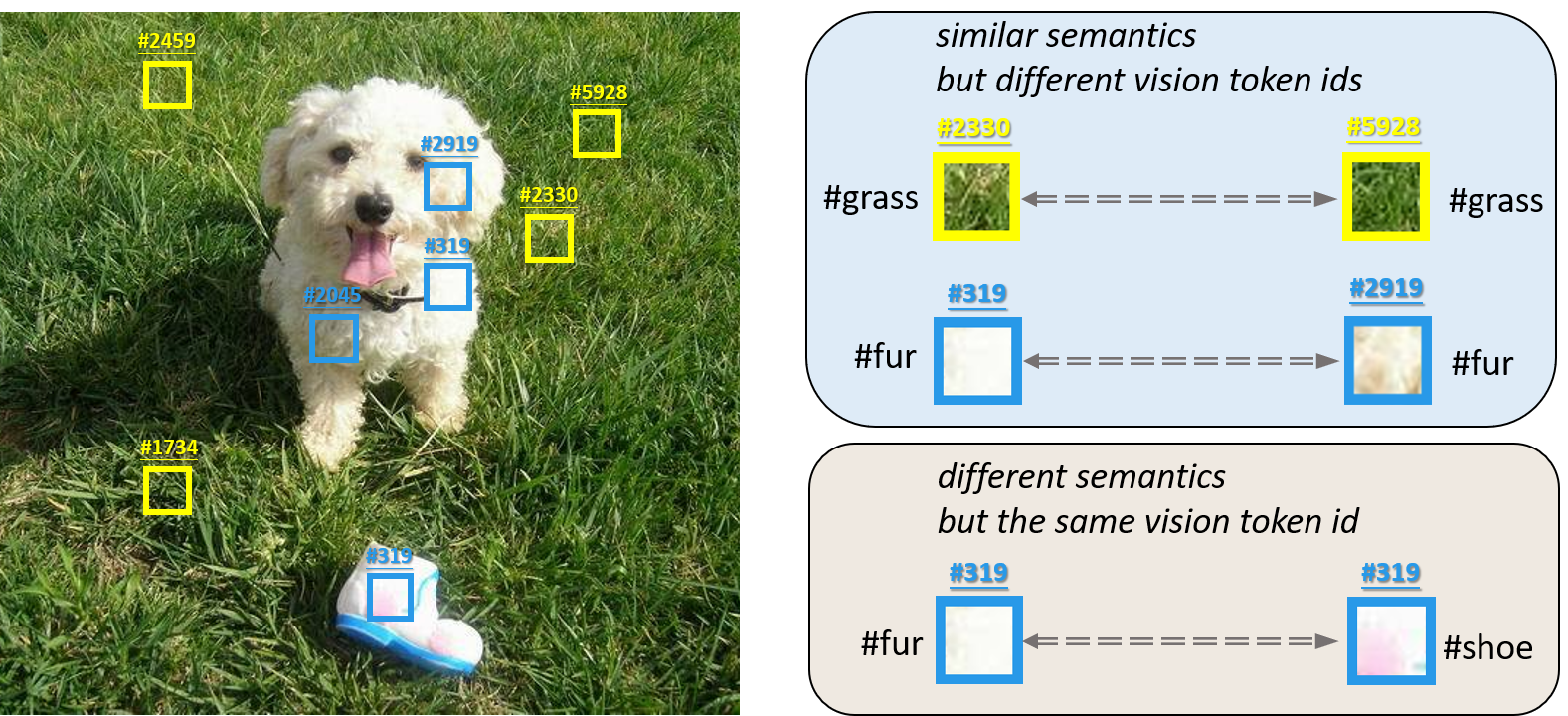}
\caption{
The improper token ids for image discretization, where a better tokenizer \cite{taming} is used here. 
% The drawbacks of purely relying on the hard discrete token label, due to the low-level training objective of tokenizer.
We observe that semantically-similar patches might be allocated with different token ids while patches with different semantics might be allocated with the same token id, indicting that the hard-label classification with unique token ids in BEiT \cite{beit} may hinder the pre-training performance. }
% We use a better tokenizer \cite{taming} which is pre-learned with both low-level reconstruction and perception-consistent objectives.}
%\vspace{-2pt}
\label{fig:drawback}
\end{center}
\end{figure}

Despite the impressive performances of BEiT on image pre-training, there remain some questions under-developed. 
(1) \textit{Does the masked patch prediction have ground-truth answers?} 
Unlike the linguistic vocabulary which is naturally composed of discrete words, the image tokenizer is relatively subjective,
% and depends on its training objective, 
\textit{i.e.}, there is no perfect answer to visual discretization and the tokenizer carries inevitable label noise even a better tokenizer is obtained in \cite{taming}.
For example, as shown in Fig. \ref{fig:drawback}, patches of the dog and the shoe are discretized into the same vision token (\#319) due to their similar pixel-level representations.
(2) \textit{Should the masked patch be assigned a unique token id given a pre-learned tokenizer?} 
Not really. 
% Is this hard label supervision good for continuous visual signals? 
As illustrated in Fig. \ref{fig:drawback}, semantically-similar patches of the grass are discretized into different vision tokens,
% due to the low-level training objective of the image tokenizer.
\textit{i.e.}, they are classified into distinct and unique ids in BEiT pre-training, neglecting their semantic relations.
% The purpose of mask prediction for self-supervised learning is to understand the image to some extent and learn meaningful semantics, instead of exactly reconstruct the visual image. 

Given the observation of the above two issues, we argue that performing MIM with a strict mapping between patch predictions and unique token ids by a hard-label classification loss in BEiT limits the visual context capturing and the pre-training performance. 
% We argue that, one feasible solution to this challenge is to ease the 
To tackle the challenge, we introduce to effectively boost BERT-style image pre-training with eased and refined masked prediction targets, that is, 
\textit{multi-choice vision token ids}.
%\textbf{multi-choice} \textbf{vision} \textbf{token} \textbf{ids}.
Rather than retraining the tokenizer with perceptual regularizations \cite{peco,ibot}, 
we efficiently inject the semantic relations into off-the-shelf vision tokens without any extra computational overhead. 
% Furthermore, as there exist no ground-truth ids for image discretization, our method is always complementary to those who explore better image tokenizers \cite{peco}.

\begin{figure}[t]
\begin{center}
\includegraphics[width=0.9\textwidth]{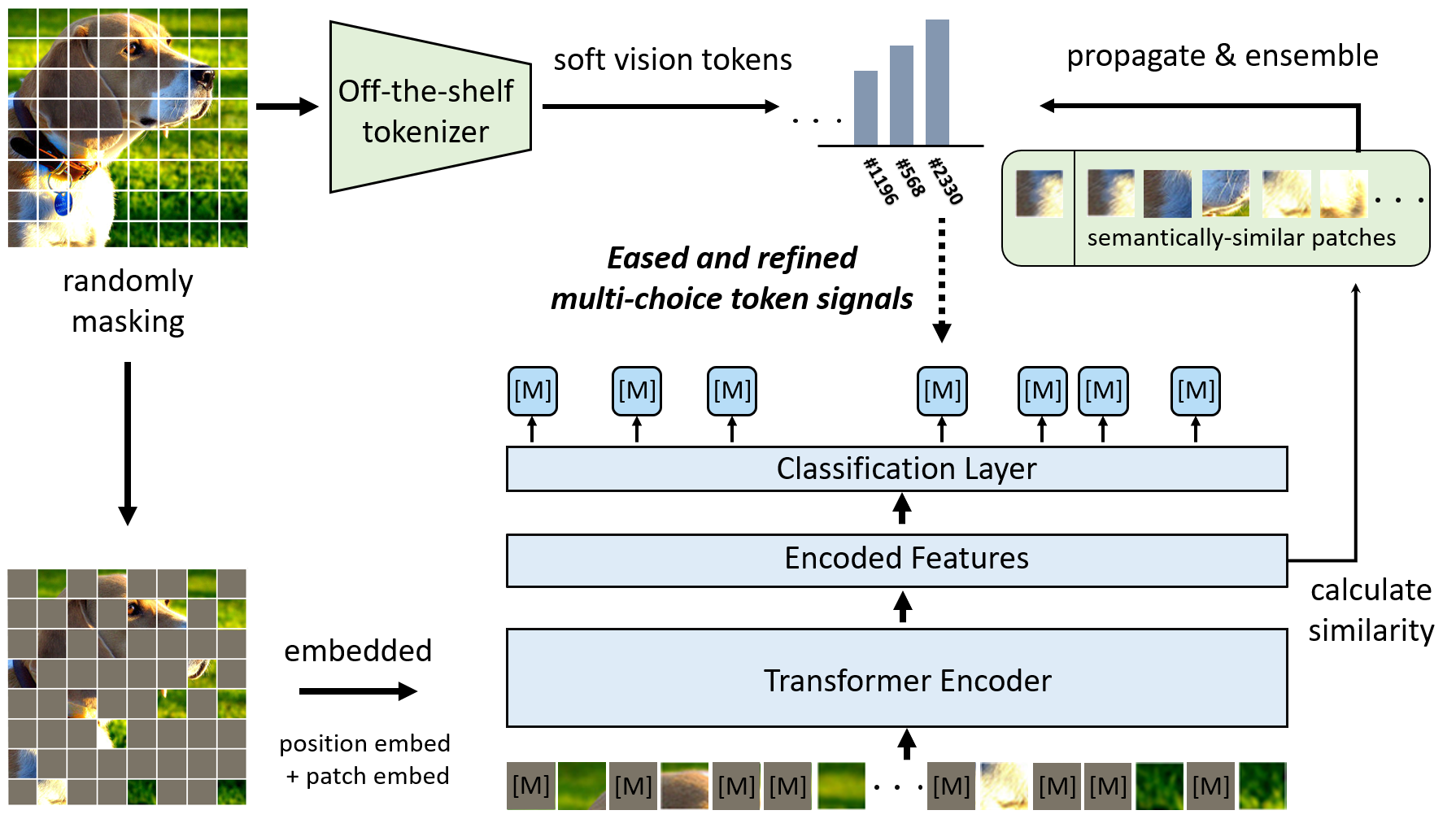}
\caption{The overview of the proposed method, \emph{mc}-BEiT. We improve image BERT pre-training with multi-choice training objectives,  which is composed of the soft probability vectors predicted by the off-the-shelf image “tokenizer”
and further refined by high-level inter-patch perceptions.
%under the observation that semantically-similar patches should share their predictions. 
A proportion of image patches are randomly masked and then fed into the vision Transformer. The masked patch prediction is optimized towards eased and refined multi-choice token ids in the form of a soft-label cross-entropy loss. }
% We minimize the soft cross-entropy loss between the masked patch prediction and the proposed eased and the proposed objective. 
% To enable the multi-choice discretization, we produce probability vectors from the off-the-shelf tokenizer and then 
% soft the output from the tokenizer as probability vectors. To enhance the global and semantic relations, we propagate the singals of semantically-similar patches to form an ensembled target. These two parts together forms an eased and refined multi-choice objective.}
% \vspace{-6pt}
\label{fig:overview}
\end{center}
\end{figure}

Specifically, to enable multi-choice answers for masked patches, we adopt the soft id probability vectors, rather than the hard predicted id over a pre-learned codebook, as the supervision signals for masked image modeling.
Although the off-the-shelf image tokenizer \cite{taming} can capture some local semantics with the training objectives of both pixel-level and perceptually-aware regularizations, it is proven to be still vulnerable to various low-level changes (see Fig. \ref{fig:drawback}). 
Therefore, we introduce to refine the predicted soft id probabilities by inter-patch semantic similarities, which are estimated by the vision Transformers being trained.
Under the observation that patches with similar high-level visual perceptions ought to share their predictions, we propagate the soft id probabilities of different patches in an image based on their semantic similarities and form ensembled learning targets for masked image patches (see Fig. \ref{fig:overview}).
The final training objective is formulated as a soft-label cross-entropy loss.

To fully evaluate our novel, flexible and effective method, we pre-train the vision Transformers with various scales on the widely-acknowledged ImageNet-1K \cite{imagenet1k} dataset and fine-tune the pre-trained models on multiple downstream tasks, including image classification, instance/semantic segmentation, and object detection. 
The empirical results show that our method impressively outperforms supervised pre-training as well as recent self-supervised learning methods \cite{mocov3,dino,beit,ibot}. 
Concretely, we achieve 84.1\% top-1 accuracy on ImageNet-1K classification with a ViT-B model, outperforming the state-of-the-art iBOT \cite{ibot} by +0.3\% with 800 fewer epochs.
% For example, a ViT-Base model pre-trained with our method achieves 84.1\% top-1 accuracy on ImageNet-1K classification. 
Regarding the transfer learning ability on different downstream tasks, our pre-trained ViT-B model achieves 49.2\% $\text{AP}^{b}$ and 44.0\% $\text{AP}^{m}$ of object detection and instance segmentation on COCO \cite{coco}, 50.8\% mIOU on downstream ADE20K \cite{ade20k} semantic segmentation, outperforming all existing methods.
%Introduction to BEiT and Problems of BEiT.
%Soft label and Structural regularization.

\section{Related Works}
Self-supervised learning (SSL) has gained great popularity benefiting from its capability of exploiting the tremendous
amounts of unlabeled data, which leverages input data itself as supervision. Substantial works \cite{moco,simclr,beit,BERT,GPT3,miles,liu2021self,cpc,decpc} have shown that the pre-training can be beneficial for downstream tasks and enable faster training convergence, which shows its impressive potentials on various machine learning tasks, especially in the fields of natural language processing and computer vision.

\subsection{BERT pre-training with masked language modeling} 
Self-supervised learning has been studied in NLP for decades. 
%Benefiting from the milestone work of Transformers \cite{attention}, the development of language pre-training has been greatly advanced. 
Masked language modeling (MLM) is firstly widely acknowledged because of BERT \cite{BERT}.
%which designs the pretext task by predicting the masked words per sentence and becomes the most famous approach due to its effectiveness and simplicity. 
BERT encourages bidirectional textual context understanding and adopts the masked language modeling approach for pre-training, which randomly masks 15\% tokens and predicts the missing words as the target. After that, various MLM variants are proposed, \textit{e.g.,} GPT \cite{GPT3}, XLM \cite{xlm}, and RoBERTa \cite{roberta}, \textit{etc}. These MLM works achieve huge success and show impressive performances on various downstream tasks, which greatly advance the development of language pre-training.

%which leverages input data itself as supervision and shows its impressive potentials in downstream tasks. 
\subsection{Self-supervised visual pre-training}
In the past few years, various pretext tasks are designed for self-supervised visual pre-training. For example, earlier pretext-based works adopt the pseudo labels based on the attributes of images to learn the representation, such as image colorization \cite{color}, jigsaw puzzle \cite{jigsaw}, context prediction \cite{doersch2015}, and rotation prediction \cite{rotation}, etc. Besides these approaches, there are two mainstream paradigms, \textit{i.e.}, contrastive learning and masked image modeling approaches, which will be further analyzed in the following subsection.

\textbf{Contrastive learning:} Contrastive learning is an instance-level discriminative approach and has occupied a dominant position in visual pre-training. Contrastive learning methods, such as SimCLR \cite{simclr,simclr_v2}, MoCo \cite{moco,mocov2,mocov3}, and Swav \cite{swav}, \textit{etc.}, typically rely on data augmentation to create the counterparts of the images and aim at learning such an embedding space, where similar sample pairs are close to each other while dissimilar ones are far apart. 
%MoCo \cite{moco} further develops the instance discrimination with momentum contrast using the query and key encoder. 
Swav \cite{swav} proposes a cluster-based contrastive learning method to enforce consistency between cluster assignments under different augmentations. 
BYOL \cite{byol} and SiamSim \cite{siamsam} abandons the negative samples and avoids the collapse with either an additional momentum network or the stop-gradient operation. 
MoCov3 \cite{mocov3} extends the contrastive learning framework for transformers and further promotes the development of self-supervised vision Transformers. 
%DINO \cite{dino} proposes a teacher-student self-distillation approach for pre-training vision Transformers.

\textbf{Masked Image Modeling:} Motivated by the great success of BERT, masked image modeling (MIM) \cite{beit,ibot,mae,conmim,maskfeat,simmim,cae} becomes a new trend in self-supervised visual pre-training, which randomly masks parts of images and reconstructs them based on the corrupted image. %iGPT \cite{igpt} firstly use pretext tasks for vision transformer pre-training. 
ViT \cite{vit} attempts to adopt masked patch prediction for self-supervised learning. 
%The reconstructed target has kinds of variants, e.g., the discrete tokens, the pixel values, and HOG features. 
BEiT \cite{beit} predicts the discrete tokens of masked token resorting to an off-the-shelf discrete VAE. Instead of discretizing the visual information, MAE \cite{mae} and SimMIM \cite{simmim} propose to directly predict the pixel-level value as the reconstruction target. MaskFeat \cite{maskfeat} further exploits different supervision signals such as HOG feature to be the objective. iBOT \cite{ibot} performs masked prediction and adopts the teacher network as an online tokenizer to provide the supervision. PeCo \cite{peco} further provides the evidence that the perceptually-aware tokenizer will provide better pre-training performance for the masked image modeling. 

\section{Preliminaries}

\subsection{Image BERT Pre-training with Masked Image Modeling}

The paradigm of mask-and-then-predict is first introduced in BERT pre-training \cite{BERT} of NLP tasks to encourage bidirectional context understanding of the textual signals. 
Recent works \cite{beit,ibot} reproduce the success of BERT by employing the proxy task of masked image modeling (MIM) on image pre-training of vision Transformers \cite{vit,deit,swinT}.
MIM requires randomly masking a proportion of the image patches and then training the vision Transformer to recover the corrupted image via reasoning among the visual context.
The pretext task of MIM enables a more fine-grained understanding of the local visual semantics compared to the contrastive counterparts \cite{simclr,moco}. 
Vision Transformers pre-trained with MIM objectives can be well transferred to a wide range of downstream tasks, \textit{i.e.}, classification, segmentation, and detection, after fine-tuning.

% Motivated by the great success of BERT, recent works \cite{beit,ibot} propose to adopt the masked image modeling (MIM) approach for pre-training Vision Transformers. MIM approaches randomly mask some portions of the images and reconstruct them as the target based on the corrupted image. The purpose of MIM is to enhance the model ability of understanding local context, and the pre-trained model can be further fine-tuned to boost the performances of various downstream tasks, \textit{i.e.,} classification, segmentation, and detection, \textit{etc.} 

% The great success of BERT in natural language pre-training motivates researcher to 

%describe the process of masking patch and then recovery. the advantage of mim: enhance local context understanding. the usage of pre-trained model, finetune for downstream task

\subsection{Masked Image Modeling as Single-choice Classification}

Introducing the mask-and-then-predict paradigm into image pre-training is actually non-trivial, because 
% the languages naturally consist of discrete words with highly abstract semantics, but 
the visual signals are continuous and cannot be predicted resorting to a well-defined vocabulary.
A pioneering work, BEiT \cite{beit}, tackles the challenge by casting masked patch prediction as a single-choice classification problem via discretizing the image into vision tokens with an off-the-shelf ``tokenizer''.
The ``tokenizer'' can be a discrete auto-encoder \cite{dalle,taming} pre-learned towards the reconstruction objective.

% and no appropriate vocabulary as in language that can be directly adopted.

% However, introducing the BERT style pre-training into vision is non-trivial because the languages are naturally consist of well-defined and highly semantic vocabulary, while the visual signal is continuous and no appropriate vocabulary as in language that can be directly adopted. To tackle this, the pioneer work BEiT \cite{beit} proposes to discretize the masked visual tokens resorting to a discrete VAE and essentially turns MIM into a classification problem.

Formally, given a raw image $x\in\mathbb{R}^{C\times H \times W}$, it is initially divided into $N$ patches $\{x_i\}_{i=1}^{N}$ and then mapped into compact patch embeddings.
We denote the corrupted image as $\hat{x}$, which is formed by masking part of the patches in $x$, and we denote the set of masked patch indices as $\mathcal{M}$.
We encode the image patch features $f(\hat{x})\in\mathbb{R}^{N\times D}$ with high-level perceptions by feeding $\hat{x}$ into the vision Transformer.
The patch features are further projected to the probabilities of the vision token ids using an MLP head which will be dropped for downstream tasks.
We denote the probability vectors as $q(f(\hat{x}))\in\mathbb{R}^{N\times V}$ where $V$ is the length of the visual vocabulary defined by the pre-learned image ``tokenizer''.

To receive the answers for the masked image modeling, we discrete the raw image $x$ into vision tokens $\{z_i\}_{i=1}^{N}$ using the image ``tokenizer'', where $z_i\in\mathbb{R}^{V}$.
The assigned token id with the maximal probability in $z_i$ is termed as $y_i$.
The pre-training objective is formulated as a hard-label cross-entropy loss to encourage masked patch prediction with unique token ids as follow,
% to minimize the hard label cross-entropy loss given the corrupted image $\tilde{x}$:
\begin{equation}
\mathcal{L}_\text{mim}(x)={\mathbb{E}_{i\in\mathcal{M}}\left[-\log q\left(y_i|f(\hat{x}_i)\right)\right]}.    
\end{equation}

% the image batch $x\in\mathrm{R}^{C\times H \times W}$ is firstly embedded into patches $x\in\mathrm{R}^{N\times (P^2 \times C)}$, where $N=HW/P^2$ and $P$ is the size of each patch. Then these image sequences $\{x_i\}_{i=1}^{N}$ are flattened into vectors and linearly projected into patch embeddings. Some portions of the images patches are randomly masked and the set of them are denoted as $\mathcal{M}$. Then the image patch features $F(x_i)$ are encoded by the vision transformer and projected to get the probabilistic predictions $\mathbb{Q}(x_i)$ using the MLP head. At the same time, the image sequences are represented as discrete tokens resorting to the pre-defined codebook. Specifically, the image $x$ is encoded by the tokenizer into logits $z=[z_1,...,z_N]\in\mathrm{R}^{N\times V}$, where $V$ is the the size of the discrete token vocabulary, and the hard vision token label is obtained as $h_i=\text{arg max}(z_i)$. Finally, the pre-training objective of BEiT is to minimize the hard label cross-entropy loss given the corrupted image $\tilde{x}$:

% \begin{equation}
% L={\mathbf{E}_{\mathcal{M}}\left[-\text{log}\,\mathbb{Q}(h_i|\hat{x_i})\right]}.    
% \end{equation}

% describe the work mechanism of BEiT, cast MIM as a classification task, using a pre-learned tokenizer to assign unique token ids for the masked patch. list the training loss (hard-label cross-entropy) here.

\section{\emph{mc}-BEiT}

BEiT provides inspiring insights of casting masked image modeling (MIM) as a classification problem to bridge the gap between discrete words in NLP tasks and continuous visual signals in computer vision tasks.
However, as there are no perfect answers for visual discretization, performing a strict mapping between patch predictions and unique token ids as a single-choice classification problem is actually a sub-optimal solution for MIM pre-training.
% Despite that BEiT pre-training can lead the vision Transformer to capture more local visual semantics, we argue that assigning image patches with unique token ids serves as a sub-optimal solution for pre-training.
As illustrated in Fig. \ref{fig:drawback}, there may exist multiple appropriate token ids for a certain patch, motivating us to boost the BEiT pre-training with multi-choice classification.
% there is no perfect answer of the visual discretization, 
% the unique vision token limits the variety of potential answers and neglects their semantic relations. 
% In fact, the purpose of mask prediction for self-supervised learning is to understand more meaningful semantics for downstream tasks, instead of exactly reconstruct the visual image. Given the observation of this, we think performing MIM with a strict mapping between patch predictions and unique token ids in the form of a
% hard-label classification loss may limit the final performance. 
%The similar patches might be allocated with different reasonable answers.

%describe your motivation and intuition xxx.

\subsection{Masked Image Modeling as Multi-choice Classification}

% To tackle this challenge, 
We introduce an improved BERT-style image pre-training with eased and refined masked prediction targets, \textit{i.e.,} {multi-choice vision token ids}, rather than a unique answer. All possible token ids in the visual vocabulary will be assigned possibilities to be chosen. To this end, we soften the training objective from the original hard-label cross-entropy loss to a soft-label cross-entropy loss with the multi-choice targets $\hat{z}\in\mathbb{R}^{N\times V}$ as follow,
\begin{equation}\label{eq:mc-mim}
\mathcal{L}_\text{mc-mim}(x)={\mathbb{E}_{i\in\mathcal{M}}\left[-\sum_{k=1}^{V}\hat{z}_{i,k} \log q\left(f(\hat{x}_{i})\right)_k\right]},
\end{equation}
where $\sum_{k=1}^{V}\hat{z}_{i,k}=1, \forall i$ and $q\left(f(\hat{x}_{i})\right)\in\mathbb{R}^{V}$.
We will go over how to produce such refined multi-choice answers for MIM pre-training in the following section.

% and we denote $\overline{\mathbb{P}}(i,k)$ as the probability to predict the vision token label of $x_i$ as $k$. 
% Thus the multi-choice objective $\overline{\mathbb{P}}\in\mathrm{R}^{N\times V}$ is used to be the optimization target for providing more diverse answers. The overall loss function is to optimize soft cross-entropy loss between the predictions and the multi-choice objective:

% \begin{equation}
% L={\mathbf{E}_{\mathcal{M}}\left[-\sum_{k=1}^{V}\overline{\mathbb{P}}(i,k)\cdot\text{log}\,\mathbb{Q}(k|\hat{x_i})\right]}.
% \end{equation}

% motivate to use soft probability vectors to form the objective of multi-choice. list the loss function (soft-label cross-entropy)

\subsection{Multi-choice Visual Discretization}

To produce multi-choice discretization without extra training stages or computational overhead, we attempt to exploit the predictions from the off-the-shelf image tokenizer.
% , which can be adopted efficiently and does not need additional training.
% To produce multi-choice discretization, we first 
Given the discretization predictions $z\in\mathbb{R}^{N\times V}$ from the image tokenizer,
we estimate the soft probabilities $p(z)\in\mathbb{R}^{N\times V}$ rather than using the unique predicted token id as done in the single-choice version.
Specifically,
% the image discretization is generally achieved by the quantized method such as Gumbel-softmax relaxation \cite{gumbel} and we can acquire the logits $z$ before the quantization. 
the soft probability vector is obtained using a softmax operation, where a temperature coefficient $\tau$ is used to move between the sharpness (single-choice) and smoothness (multi-choice),
% control the smoothness of the probabilistic distribution:
\begin{equation}
p(z_i)_k = \frac{\text{exp}(z_{i,k}/\tau)}{\sum_{j=1}^{V}\text{exp}(z_{i,j}/\tau)}.
\end{equation}

% \begin{equation}
% \mathbb{P}(k) = \frac{\text{exp}(z_k/\tau_p)}{\sum_{i=1}^{V}\text{exp}(z_i/\tau_p)},
% \end{equation}

As discussed in the introduction section and illustrated in Fig \ref{fig:drawback}, semantically-similar patches may be allocated with discrepant token ids and semantically-dissimilar patches may be allocated with the same token id due to their low-level similarities, indicating that the raw predictions from the off-the-shelf tokenizer are sub-optimal to fully represent the semantic relations among patches. 
The phenomenon motivates us to refine the predictions of the tokenizer with inter-patch relations.
% and further enhance the inter-patch semantic relations. 
%the signals obtained from the tokenizer neglect the similar semantics of patches, which motivates us to refine the signals and enhance the inter-patch semantic relation. 
% We propose to bridge the low-level semantics and high-level perceptions with the help of semantically-similar patches.
The inter-patch relations can be estimated with their high-level perceptions, which are encoded by the in-training vision Transformer.
Specifically, we calculate the cosine similarity between patch features to measure their affinities,
\begin{equation}
    W(\hat{x}_i)_k=\frac{\text{exp}{\langle f(\hat{x}_i)}, {f(\hat{x}_k)}\rangle}{\sum_{j=1}^N{\text{exp}\langle{f(\hat{x}_i)}, {f(\hat{x}_j)}\rangle}},
\end{equation}
where $W(\hat{x})\in\mathbb{R}^{N\times N}$ and $\langle\cdot,\cdot\rangle$ indicates the inner product between two feature vectors after $\ell_2$ normalization.
% The encoded features can be viewed as the semantic representation of the patch and the dot product of the encoded features is adopted to estimate the inter-patch similarity $W\in\mathbb{R}^{N\times N}$, where $f(\hat{x}_i)$ denotes the $l_2$ normalized encoded feature of the image patch $\hat{x}_i$:
% \begin{equation}
%     \text{W}(i,j)=\frac{\text{exp}(\sigma({F(x^i)})^T \sigma({F(x^j)}))}{\sum_{k=1}^N{\text{exp}(\sigma({F(x^i)})^T \sigma({F(x^k)}))}}.
% \end{equation}
% \begin{equation}
% s_{i,j}=\sigma({F(x^i)})^T \sigma({F(x^j)}),
% \end{equation}
Based on the observation that perceptually-similar patches ought to share their choices, we propagate the soft probabilities $p$ of different patches in an image $x$ to form a refined target $W(\hat{x}) p(z)\in\mathbb{R}^{N\times V}$. In this way, patches with similar high-level perceptions can provide complementary supervision signals for the masked patches.
% based on the inter-patch similarity.

% \begin{equation}
%     \text{W}(i,j)=\frac{\text{exp}(s_{i,j})}{\sum_{k=1}^N{\text{exp}(s_{i,k})}}.
% \end{equation}

The overall objective of multi-choice image discretization is composed of the weighted sum of the aforementioned parts, where the semantic equilibrium coefficient $\omega$ is introduced to move between low-level semantics (directly predicted by the tokenizer) and high-level semantics (ensembled from the perceptually-similar patches).
% avoid collapse due to the inevitable noise to calculate the patch similarity, especially in the early epochs. 
The former one adopts the eased supervision directly predicted from the tokenizer, while the latter one injects high-level perceptions by propagating among other semantically-similar patches, together forming the refined multi-choice targets $\hat{z}\in\mathbb{R}^{N\times V}$ as follow:

% \begin{equation}
%     L=-\mathbf{E}[(\omega \mathbf{W} \mathbf{P}_{\bar{\Omega}} + (1-\omega)\mathbf{P}_{\Omega})\text{log}\mathbf{Q}]
% \end{equation}

\begin{equation}
\hat{z}=\omega p(z) + (1-\omega)W(\hat{x})p(z),
\end{equation}
which is further used as the objectives for masked patch predictions in Eq. (\ref{eq:mc-mim}).

\section{Experiments}

% \subsection{The compared methods}
% In the experiments, we adopt the five related works as the compared methods.  
\subsection{Pre-Training Setup}

In our experiments, the images of 224$\times$224 resolution are divided into 14$\times$14 image sequences with 16$\times$16 patch size. We use different architectures such as ViT-Base/16 and ViT-Large/16 for pre-training and the backbone implementation follows \cite{vit} for fair comparisons. 
%The relative position embedding is adopted the default protocol without specific notification. 
For the BERT-style visual pre-training, we randomly mask 75\% patches for masked image modeling. Inspired by PeCo \cite{peco}, we employ the off-the-shelf VQGAN of \cite{taming} as a better tokenizer, which is pre-trained on OpenImages \cite{OpenImages2} with the vocabulary size of 8192. In our experiments, the semantic equilibrium coefficient $\omega$ is 0.8 and the temperature coefficient $\tau$ is 4.0 by default. The vision Transformers are pre-trained for 800 epochs on the widely-acknowledged ImageNet-1K \cite{imagenet1k} dataset, which includes 1.28 million images. Note that the ground-truth labels are disabled for pre-training. We use 16 Nvidia A100 GPUs for pre-training and a batch size of 128 per GPU. We adopt simple image augmentation for pre-training, including random resized cropping and horizontal flipping. 
The detailed recipe of pre-training and finetuning is summarized in the Appendix.

\begin{table}[t]

\small
\centering
\caption{The top-1 fine-tuning accuracy of ImageNet-1K using ViT-Base and ViT-Large with different pre-training methods.}
\label{table:imagenet1k}
%\resizebox{\textwidth}{!}{
\setlength{\tabcolsep}{6mm}{
%\vspace{-3pt}
\begin{tabular}{llcc}
%\hline
\toprule[1pt]
\multicolumn{1}{l}{Method} & Reference & Pre-train Epoch & Acc (\%)  \\ \hline
\multicolumn{4}{l}{\textsl{Supervised Pre-training (training from scratch):}}\\ \hline
ViT-B/16 \cite{vit}  & ICLR 2021 & - & 77.9 \\
ViT-L/16 \cite{vit}  & ICLR 2021 & - & 76.5 \\
DeiT-B/16 \cite{deit}  & ICML 2021 & - & 81.8 \\
 \hline
\multicolumn{4}{l}{\textsl{Self-supervised Pre-training using} ViT-B/16:}\\ \hline
MoCo v3 \cite{mocov3} & CVPR 2021 & 300 & 83.2 \\
DINO \cite{dino} & ICCV 2021 & 300 & 82.8 \\
%Ours & this paper & 300 & \bf{83.9} \\
BEiT \cite{beit} & ICLR 2022 & 800  & 83.2 \\
iBOT \cite{ibot} & ICLR 2022 & 1600 & 83.8 \\
MAE \cite{mae} & CVPR 2022 & 1600  & 83.6 \\
SimMIM \cite{simmim} & CVPR 2022 &  800 & 83.8 \\

Ours & this paper & 800 & \bf{84.1} \\ \hline

\multicolumn{4}{l}{\textsl{Self-supervised Pre-training using} ViT-L/16:}\\ \hline
MoCo v3 \cite{mocov3} & CVPR 2021 & 300 & 84.1 \\
%DINO \cite{dino} & ICCV 2021 & - &  -\\
BEiT \cite{beit} & ICLR 2022 & 800 & 85.2 \\
%iBOT \cite{ibot} & ICLR 2022 & - &  \\
MAE \cite{mae} & CVPR 2022 & 1600  & \textbf{85.9} \\
Ours & this paper & 800 & 85.6 \\

%\hline
\bottomrule[1pt]
\end{tabular}
}
%}
%\vspace{-0.5cm}
\end{table}

\subsection{Image Classification}
% The vision transformers are pre-trained on the widely-acknowledged ImageNet-1K dataset and Top-1 accuray is adopted as the evaluation metric after fine-tuning. For the ImageNet classification task, the fully connected layer is employed for the classification task after the average pooling of the feature embeddings.

For the ImageNet classification task, the fully-connected layer is employed as the classifier after the average pooling of the feature embeddings. We adopt top-1 accuracy after fine-tuning as the evaluation metric and we thoroughly compare our method with the supervised methods, \textit{i.e.,} ViT \cite{vit}, DeiT \cite{deit}, and recently published state-of-the-art self-supervised learning methods, \textit{i.e.,} MoCo v3 \cite{mocov3}, DINO \cite{dino}, BEiT \cite{beit}, and iBOT \cite{ibot}. Besides, we also compare with the very recent pixel-level MIM methods, \textit{i.e.,} MAE \cite{mae} and SimMIM \cite{simmim}. The experiment results are listed in Tab. \ref{table:imagenet1k}. As observed from the results, the proposed method obtains 84.1\% top-1 accuracy on ViT-B, outperforming the competing methods and achieving state-of-the-art performance. We can see that our \emph{mc}-BEiT shows significant gains compared to the baseline BEiT, which verifies the effectiveness of our introduced multi-choice objectives. Concretely, our method outperforms the recent state-of-the-art method iBOT \cite{ibot} by +0.3\% with the fewer 800 epochs pre-training. It is noted that iBOT adopts an extra teacher network and enables multi-crops for pre-training, showing lower efficiency than our method.
% therefore our method is also both effective and more efficient. 

\begin{table}[h]

\small
\centering
\caption{The top-1 fine-tuning accuracy of ImageNet-1K using our \emph{mc}-BEiT with different training epochs and backbone architectures.}
\label{table:imagenet_epochs}
%\resizebox{\textwidth}{!}{
\setlength{\tabcolsep}{5mm}{
\begin{tabular}{lcccc}
%\hline
\toprule[1pt]
\multicolumn{1}{c}{Method} & Arch. & Model Size& Pre-train Epoch & Acc (\%)  \\ \hline
\multicolumn{4}{l}{\textsl{Self-supervised Pre-training using} ViT-B/16:}\\ \hline
%Ours & this paper & 300 & \bf{83.9} \\
Ours & ViT-B & 86M & 100 & 83.3 \\ 
Ours & ViT-B & 86M & 300 & 83.9 \\ 
Ours & ViT-B & 86M & 800 & \textbf{84.1} \\ \hline
\multicolumn{4}{l}{\textsl{Self-supervised Pre-training using} ViT-L/16:}\\ \hline
Ours & ViT-L & 307M & 300 & 85.2 \\
Ours & ViT-L & 307M & 800 & \textbf{85.6} \\
%\hline
\bottomrule[1pt]
\end{tabular}
}
%}
%\vspace{-0.5cm}
\end{table}

\noindent\textbf{Different training epochs and architectures:} We also provide more comprehensive results of different training epochs and architectures in Tab. \ref{table:imagenet_epochs}. From the table, we can see that our method can adapt well to different scales of vision tranformers, \textit{e.g.,} the mostly used ViT-B and ViT-L. It is worth noting that our method obtains a relatively high accuracy (already achieves the state-of-the-art performance) when pre-training for only 300 epochs. Moreover, the performance can be further improved with longer pre-training epochs, \textit{e.g.}, the accuracy reaches 84.1\% pre-training for 800 epochs.

\begin{figure}[t]
\begin{center}
\includegraphics[width=0.7\textwidth]{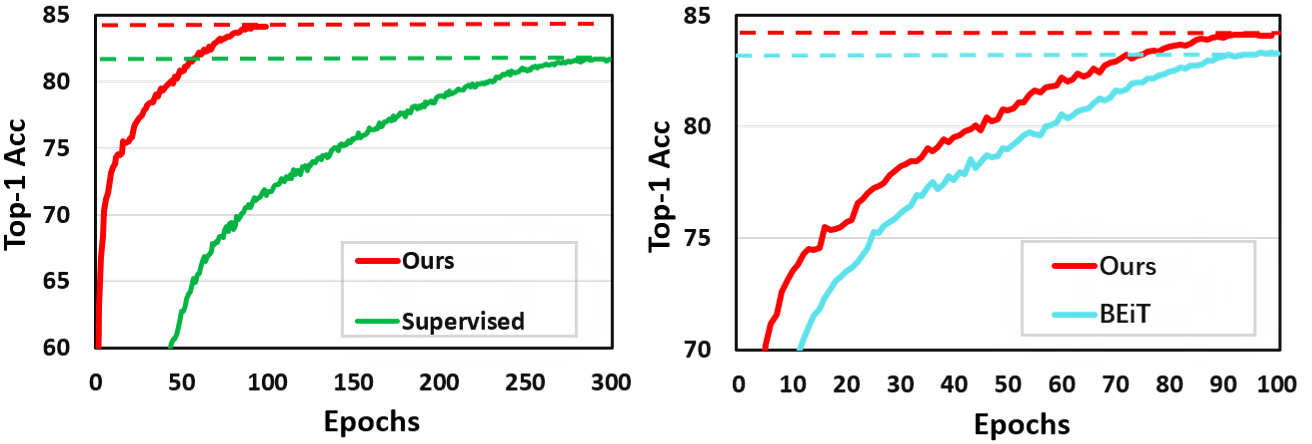}
\caption{The convergence curves when fine-tuning ViT-B models on ImageNet-1K classification. The models are pre-trained by different methods.}
% \vspace{-3pt}
\label{fig:conver}
\end{center}
\end{figure}

\noindent\textbf{Convergence curve:} In Fig \ref{fig:conver}, we further demonstrate the convergence curve of the supervised learning method and self-supervised learning methods, \textit{i.e.,} the baseline BEiT and our method, when fine-tuning ViT-B models. As shown in the figure, the proposed method achieves faster convergence as well as better performance than training DeiT from scratch \cite{deit}. Meanwhile, our method obtains obvious and consistent performance gains compared to the baseline method BEiT, showing the superiority of the proposed multi-choice training objectives.

\subsection{Object Detection and Instance Segmentation}
For object detection and instance segmentation tasks, COCO \cite{coco} benchmark is employed to validate the pre-training performances. 
%Here, we use absolute position embedding and interpolate it to adapt to the multi-scale strategy. 
We follow the implementation of \cite{benchmarking,mimdet} and the model is trained for 25 epochs (we also provide another evaluation setting following iBOT \cite{ibot} in the Appendix).
ViT-B is adopted as the backbone and Mask-RCNN \cite{maskrcnn} is used as the task head. The evaluation metrics for objection detection and instance segmentation are bounding box AP and mask AP, respectively.

\begin{table}[h]
\small
\centering
\caption{Experiment results of object detection and instance segmentation on COCO. We follow the implementation of \cite{benchmarking,mimdet} and the model is trained for 25 epochs. Intermediate fine-tuning denotes the model is further fine-tuned on ImageNet-1K. 
% Mask R-CNN is adopted and the model is trained for 25 epochs. 
% The experiments setting and training recipe follow the implementation of \cite{benchmarking}.
%As experiments on COCO are not conducted in BEiT, the results are based on our re-implementation.
}
\label{coco}
%\resizebox{\textwidth}{!}{
\setlength{\tabcolsep}{3.5mm}{
\begin{tabular}{l|lcc}
%\hline
\toprule[1pt]
\multicolumn{1}{l|}{\multirow{2}{*}{Method}} &
  \multicolumn{1}{l}{\multirow{2}{*}{Reference}} &
  \multicolumn{1}{c}{Object Det.} &
  Instance Seg. \\ %\cline{3-4} 
\multicolumn{1}{l|}{} &
  \multicolumn{1}{c}{} &
  \multicolumn{1}{c}{$\text{AP}^{b}$} &
  \multicolumn{1}{c}{$\text{AP}^{m}$} \\ \hline
% \multicolumn{1}{l|}{Method} & Reference & $\text{AP}^{b}$ & $\text{AP}^{m}$ \\ \hline
Supervised \cite{deit} & ICML 2021 & 46.5 & 41.7\\
MoCo v3 \cite{mocov3} & CVPR 2021 & 46.6 & 41.9 \\
DINO \cite{dino} & ICCV 2021  & 47.6 & 42.3 \\
MAE \cite{mae} & CVPR 2022  & 48.0 & 43.0 \\
iBOT \cite{ibot} & ICLR 2022 & 48.4 & 42.9 \\
BEiT \cite{beit} & ICLR 2022 & 47.6 & 42.2 \\
Ours & this paper & 48.5 & 43.1 \\ \hline
+Intermediate Fine-tuning \\ \hline
BEiT \cite{beit} & ICLR 2022 & 48.4 & 43.5 \\
Ours & this paper & \textbf{49.2} & \textbf{44.0} \\

%\hline
\bottomrule[1pt]
\end{tabular}
}
%}
\end{table}

% \begin{table}[t]
% \setlength{\abovecaptionskip}{0.cm}
% \small
% \centering
% \caption{Experiment results of object detection and instance segmentation on COCO. Intermediate fine-tuning denotes the model is further fine-tuned on ImageNet-1K. Cascaded Mask R-CNN and 1$\times$ training schedule are adopted. As experiments on COCO are not conducted in BEiT, the results are based on our re-implementation.}
% \label{coco}
% %\resizebox{\textwidth}{!}{
% \setlength{\tabcolsep}{3.5mm}{
% \begin{tabular}{l|ccc}
% %\hline
% \toprule[1pt]
% \multicolumn{1}{l|}{\multirow{2}{*}{Method}} &
%   \multicolumn{1}{c}{\multirow{2}{*}{Reference}} &
%   \multicolumn{1}{c}{Object Det.} &
%   Instance Seg. \\ %\cline{3-4} 
% \multicolumn{1}{l|}{} &
%   \multicolumn{1}{c}{} &
%   \multicolumn{1}{c}{$\text{AP}^{b}$} &
%   \multicolumn{1}{c}{$\text{AP}^{m}$} \\ \hline
% % \multicolumn{1}{l|}{Method} & Reference & $\text{AP}^{b}$ & $\text{AP}^{m}$ \\ \hline
% Supervised \cite{deit} & ICML 2021 & 47.9 & 42.9\\
% MoCo v3 \cite{mocov3} & CVPR 2021 & 47.9 & 42.7 \\
% DINO \cite{dino} & ICCV 2021  & 50.1 & 43.4 \\
% iBOT \cite{ibot} & ICLR 2022 & 51.2 & 44.2 \\
% BEiT \cite{beit} & ICLR 2022 & 49.6 & 42.8 \\
% Ours & this paper & 50.1 & 43.1 \\ \hline
% +Intermediate Fine-tuning \\ \hline
% BEiT \cite{beit} & ICLR 2022 & 50.7 & 43.8 \\
% Ours & this paper & \textbf{51.2} & \textbf{44.3} \\

% %\hline
% \bottomrule[1pt]
% \end{tabular}
% }
% %}
% \end{table}
As observed in Tab. \ref{coco}, the BERT style pre-training shows superiority to supervised pre-training in terms of performances. Our method achieves 48.5\% and 43.1\% in $\text{AP}^{b}$ and $\text{AP}^{m}$. Meanwhile, the proposed method outperforms the competitor BEiT with +0.9\%/+0.9\%  gain in $\text{AP}^{b}$ and $\text{AP}^{m}$. We also evaluate the performance after intermediate fine-tuning, the relative improvement is still obvious, \textit{i.e.,} +0.8\%/+0.5\% to BEiT. Our method even outperforms the recent pixel-level MIM method MAE and obtains better performances compared to state-of-the-art methods.

\subsection{Semantic Segmentation}

Semantic segmentation belongs to the pixel-level classification task and is often adopted to evaluate the pre-training performance on downstream tasks. Here we evaluate the performance on ADE20k \cite{ade20k} benchmark and mean intersection over union (mIOU) averaged over all semantic categories is adopted as the evaluation metric. Following the common setting in \cite{beit,ibot}, ViT-B is adopted as the default backbone and UPerNet \cite{upernet} is used for semantic segmentation task head.

\begin{table}[t]

\small
\centering
\caption{Results of semantic segmentation on ADE20K. Intermediate fine-tuning denotes the pre-trained model has been fine-tuned on ImageNet-1K classification. 
%The vision Transformer backbone is ViT-B and the task head is UPerNet.
}
\label{table:ade20k}
%\resizebox{\textwidth}{!}{
\setlength{\tabcolsep}{7mm}{
\begin{tabular}{l|lc}
%\hline
\toprule[1pt]
\multicolumn{1}{l|}{Method} & Reference & mIOU  \\ \hline
Supervised \cite{deit} & ICML 2021 & 45.3 \\
MoCo v3 \cite{mocov3} & CVPR 2021 & 47.2 \\
DINO \cite{dino} & ICCV 2021  & 46.8 \\
MAE \cite{mae} & CVPR 2022 & 48.1 \\   
iBOT \cite{ibot} & ICLR 2022 & 50.0 \\ 
BEiT \cite{beit} & ICLR 2022 & 45.6 \\
Ours & this paper & 47.0 \\
\hline
+Intermediate Fine-tuning \\ \hline
BEiT \cite{beit} & ICLR 2022 & 47.7 \\
Ours & this paper & \textbf{50.8} \\

%\hline
\bottomrule[1pt]
\end{tabular}
}
%}
%\vspace{-0.5cm}
\end{table}

Because the pre-training process does not introduce the instance discrimination, the performance can be further improved after intermediate fine-tuning on ImageNet-1K according to BEiT \cite{beit}. Therefore we also compare the performance after intermediate fine-tuning. Tab. \ref{table:ade20k} shows that our method significantly improves the transferability of pre-trained models compared to the supervised learning, with +5.5\% performance gain. It is also noticed that our method outperforms recent state-of-the-art self-supervised methods. It achieves better results as 47.0\%/50.8\% mIOU and improves +1.4\%/3.1\% gain to its pre-training only version.

\section{Ablation Study}
In this section, we conduct an extensive ablation study of our method on ImageNet-1K. Considering the time expenditure, all ablation experiments are performed under 100-epoch pre-trained ViT-B/16 on ImageNet-1K.

\subsection{The temperature coefficient $\tau$}
The hyper-parameter of temperature coefficient $\tau$ is to scale the logits from the tokenizer, which moves between the sharpness (single-choice) and smoothness (multi-choice). We adopt the common values for temperature to ablate its effect. In general, the small temperature will sharp the probability distribution and the large one will smooth it conversely. When $\tau$ is extremely small, it is an approximate single-choice classification task. The ablation is shown in the Fig. \ref{fig:hyper}(a), where \textit{single-choice label} indicates training with the strict mapping to the unique answer. From the result, we can observe that multi-choice vision token improves the BERT style pre-training performance and it behaves better when setting temperature factor at 4.0 empirically.

% \begin{table}[h]
% \small
% \centering
% \caption{Ablation study on the hyper-parameter of temperature coefficient $\tau$.}
% \label{temp}
% %\resizebox{\textwidth}{!}{
% \setlength{\tabcolsep}{5mm}{
% % \begin{tabular}{c|c|cccc}
% % %\hline
% % \toprule[1pt]
% \begin{tabular}{lc|cccc}
% \toprule[1pt]
% \multicolumn{1}{l|}{} & \multicolumn{1}{c|}{\multirow{2}{*}{Single-choice Label}} & \multicolumn{4}{c}{Multi-Choice Label}                                                 \\ \cline{3-6} 
% \multicolumn{1}{l|}{} & \multicolumn{1}{c|}{}                            & \multicolumn{1}{c}{0.04} & \multicolumn{1}{c}{1.0} & \multicolumn{1}{l}{4.0} & 10.0 \\ \hline
% % \multicolumn{1}{c|}{} & Hard Label & 0.04 & 1.0 & 4.0 & 10.0\\ \hline
% \multicolumn{1}{c|}{Acc} & 83.0 & 83.1 & 83.1&  \textbf{83.3} & 83.2  \\

% %\hline
% \bottomrule[1pt]
% \end{tabular}
% }

%}
% \end{table}

\subsection{The semantic equilibrium coefficient $\omega$}

\begin{figure}[H]
% \setlength{\abovecaptionskip}{0.cm}
% \setlength{\belowcaptionskip}{-0.cm}
%\vspace{-3pt}
\begin{center}
\includegraphics[width=0.8\textwidth]{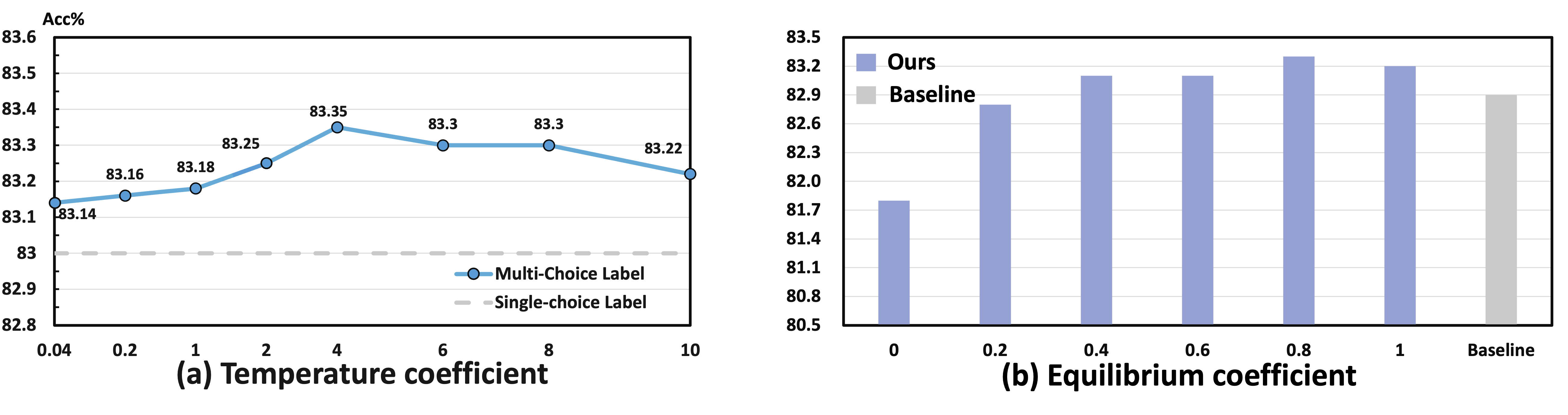}
\caption{Ablation study on the hyper-parameters.}
% \vspace{-5pt}
\label{fig:hyper}
\end{center}
\end{figure}

%The semantic equilibrium coefficient $\omega$ is used to trade off the strength of the soft label produced by tokenizer and the refined label ensembled from the semantically-similar patches. 
The semantic equilibrium coefficient $\omega$ is introduced to move between low-level semantics (directly predicted by the tokenizer) and high-level semantics (ensembled from the perceptually-similar patches).
The ablation study is shown in Fig. \ref{fig:hyper}(b). When setting $\omega$ to 0, the objective relies on totally the inter-relationship guided objective and it achieves only 81.8\% accuracy, which is because the inevitable noise of calculating patch similarity, especially in the early epochs, will cause collapse and degrade the pre-learning performance. As the coefficient goes larger, it shows consistent gains over baseline. When setting $\omega$ to 1.0, the objective comes only from the low-level signals of the tokenizer and the performance is still higher than baseline, which shows the superiority of multi-choice to single-choice. As observed from the results, the semantic equilibrium coefficient is thus set to be 0.8 for better performances.

% \begin{table}[h]
% \setlength{\abovecaptionskip}{0.cm}
% \setlength{\belowcaptionskip}{-0.cm}
% \small
% \centering
% \caption{Ablation study on the trade-off parameter.}
% \label{table:weight}
% %\resizebox{\textwidth}{!}{
% \setlength{\tabcolsep}{5mm}{
% \begin{tabular}{c|cccccc}
% %\hline
% \toprule[1pt]
% \multicolumn{1}{c|}{} & 0 & 0.2 & 0.4 & 0.6 & 0.8 & 1.0 \\ \hline
% \multicolumn{1}{c|}{Top-1 Acc} & 83.2 & 83.3 & 83.1& 83.1& 82.8 &  81.8 \\

% %\hline
% \bottomrule[1pt]
% \end{tabular}
% }

% %}
% \vspace{-0.5cm}
% \end{table}

\subsection{Masking strategy}
In the masked image modeling approach, the masking strategy determines the difficulty of inferring the missing patches. Tab. 3 shows the influence of different mask strategies, where \textit{Block} and \textit{Random} masking types and different mask ratios are conducted for ablation. It is observed from the results that the random masking strategy with 75\% masking ratio makes the best performances, which is thus adopted as the default setting for pre-training.

% \begin{figure}[H]
% % \setlength{\abovecaptionskip}{0.cm}
% % \setlength{\belowcaptionskip}{-0.cm}
% %\vspace{-3pt}
% \begin{center}
% \includegraphics[width=1.0\textwidth]{mask.png}
% \caption{Ablation study on different masking strategies.}
% \label{fig:mask}
% \end{center}
% \end{figure}

\begin{table}[t]
\small
\centering
% \vspace{-3pt}
\caption{Ablation study on different masking strategies.}
\label{table:pacs}
%\resizebox{\textwidth}{!}{
\setlength{\tabcolsep}{7mm}{
\begin{tabular}{c|cc}
%\hline
\toprule[1pt]
\multicolumn{1}{c|}{Masking Strategy} & Masking Ratio & Acc (\%)  \\ \hline
Block & 45 \% & 83.2 \\
Block & 60 \% & 83.2 \\
Block & 75 \% & 82.8 \\
\hline
Random & 45 \% & 83.0 \\
Random & 60 \% & 83.1 \\
Random & 75 \% & \textbf{83.3} \\
Random & 90 \% & 83.0 \\

%\hline
\bottomrule[1pt]
\end{tabular}
}

%}
\end{table}

\subsection{Tokenizer}

In the BERT-style visual pre-training, the tokenizer plays the role of a vocabulary in texts and is used to produce the discrete vision token as supervision. As discussed in PeCo \cite{peco}, perceptually-aware tokenizer may benefit the image BERT pre-training, so we introduce to use off-the-shelf VQGAN \cite{taming} as a better tokenizer throughout our experiments. Besides, we would like to also verify the effectiveness of our multi-choice objectives on top of the vanilla BEiT.
%BEiT adopts the discrete VAE in DALL-E \cite{dalle} as the tokenizer, which utilizes over 250 million images for training. As discussed in iBOT \cite{ibot} and PeCo \cite{peco}, the capability of tokenizer to capture high-level semantics is important to the performance of BERT-style pre-training. Hence, we conduct the experiments on different dVAE tokenizer. VQGAN is trained to learn perceptual discrete vision tokens and thus is conducted to verify the observation. Instead of retraining the tokenizer, we directly borrow the pre-trained model in \cite{taming} for experiments. 

\begin{table}[h]
\small
\centering
\caption{Ablation study on the different tokenizer.}
\label{tokenizer}
%\resizebox{\textwidth}{!}{
\setlength{\tabcolsep}{4mm}{
\begin{tabular}{c|c|c|c|c}
%\hline
\toprule[1pt]

% \multicolumn{1}{l|}{} & \multicolumn{1}{c|}{\multirow{2}{*}{Source}} & \multicolumn{2}{c}{Top 1 Acc.}                  \\ \cline{3-4} 
% \multicolumn{1}{l|}{} & \multicolumn{1}{c|}{}                        & Baseline             & \multicolumn{1}{c}{Ours} \\ \hline
\multicolumn{1}{l|}{} & \multicolumn{2}{c|}{Training Data}                       & \multicolumn{2}{c}{Top 1 Acc. (100 / 800 epochs)}                  \\ \cline{2-5} 
\multicolumn{1}{l|}{} & \multicolumn{1}{c|}{Source} & \multicolumn{1}{c|}{Scale} & BEiT             & \multicolumn{1}{c}{Ours} \\ \hline
\multicolumn{1}{c|}{DALL-E \cite{dalle}} & Private &  250M & 82.3 / 83.2 & 82.6 / 83.7 \\
\multicolumn{1}{c|}{VQGAN \cite{taming}} & OpenImage &  9M & 82.9 / 83.8 & 83.3 / 84.1\\

%\hline
\bottomrule[1pt]
\end{tabular}
}

%}
\end{table}

The influence of tokenizers is shown in Tab. \ref{tokenizer}. It is shown that adopting the VQGAN as the tokenizer brings better performance than DALL-E, which verify our observation that tokenizer with high semantics can indeed improve the pre-training performance. It also indicates that enhancing the semantic relation is beneficial to visual pre-training. Meanwhile, it is noticed that the relative improvement of our method is consistent regardless of different kinds of tokenizers, which demonstrates the effectiveness of the proposed method.

\subsection{Visualization}

\begin{figure}[t]
\begin{center}
\includegraphics[width=0.75\textwidth]{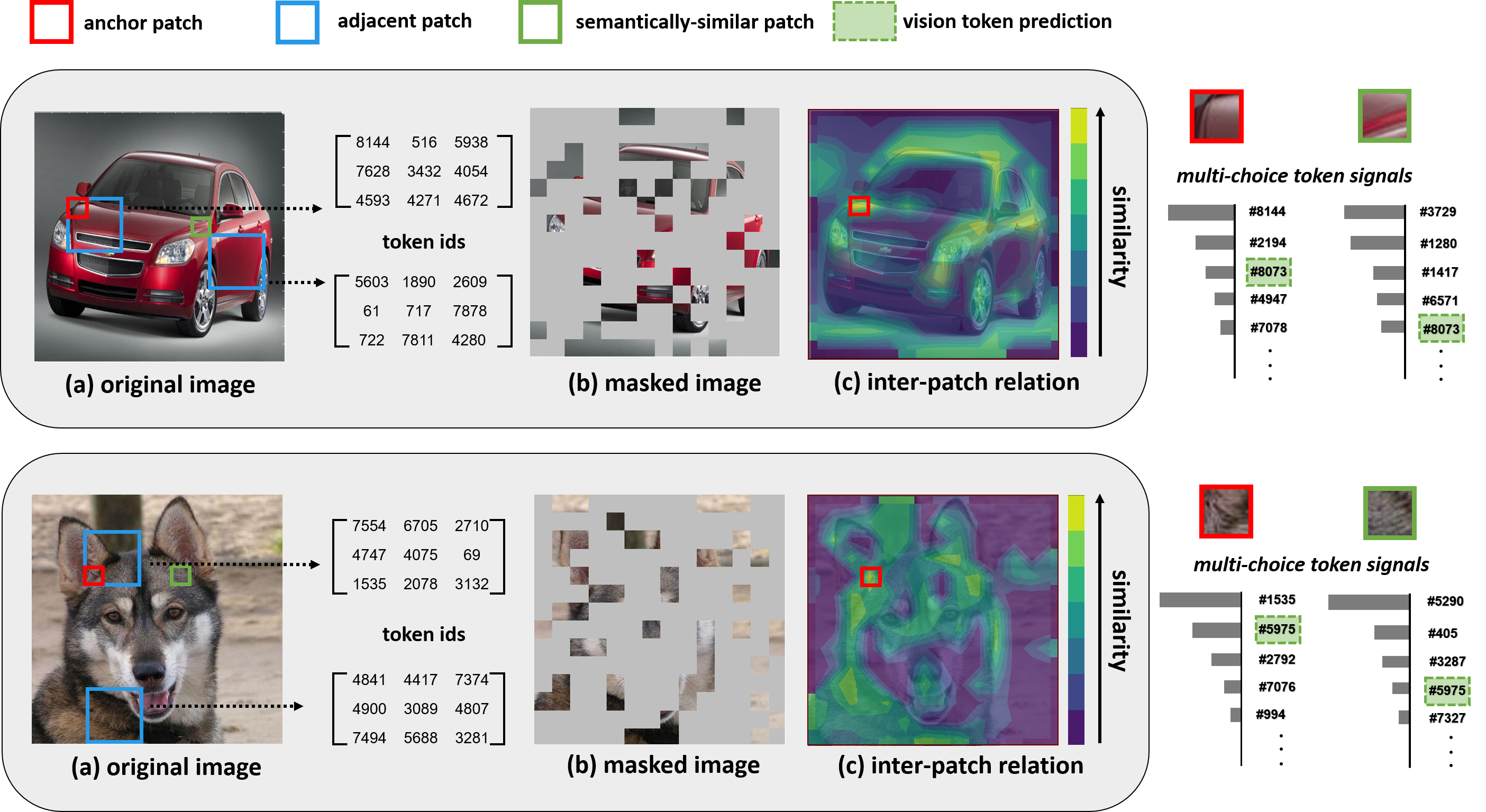}
\caption{The visualization is obtained using the off-the-shelf tokenizer and our pre-trained vision Transformer. The inter-patch perception relation is equipped with contour lines for better visual effect.}
% \caption{The visualization is obtained using the off-the-shelf tokenizer and our pre-trained vision Transformer. (a): the original image. (b) corrupted image under 75\% random masking. (c): inter-patch perception relation, which is equipped with contour lines for better visual effect.}
% \vspace{-2pt}
\label{vis}
\end{center}
\end{figure}

Besides the quantitative experiment results, we further provide some visualizations in Fig. \ref{vis} for better understanding the effects of our multi-choice answers. 75\% patches of the images are randomly masked for prediction. 

It can be observed from the blue box in Fig. \ref{vis}(a), the adjacent patches with similar semantics are still allocated with different vision token ids, indicating that the hard vision token id directly from the tokenizer neglects the semantic relations and is a sub-optimal objective. In contrast, the proposed eased and refined objective can provide diverse possible vision tokens for the prediction. As shown in our multi-choice token signals, the semantically-similar patches have the possibility to be allocated with the same vision token, which refines the objective with inter-patch perceptions. Furthermore, we randomly select a masked patch and shows the inter-patch perception relations (obtained from the patch feature similarity) learned by the pre-trained model in Fig. \ref{vis}(c). 
%Taking the patch located at the edge of the car for illustration, 
The similar patches can still be well estimated even under heavy random masking and the inter-patch relation shows higher responses, \textit{e.g,} the skeleton of the car. 
It demonstrates that the informative semantic relations estimated by the in-training vision Transformer can properly enhance the multi-choice discretization for pre-training.
%It demonstrates that the pre-trained model can well capture and predict the semantics of the missing patches, showing our effectiveness to enhance the model's ability to capture semantic relations.
%Benefiting from the multi-choice answers, the semantically-similar patches can share their predictions and enhance the model's ability to capture high-level semantics.

%To verify our observation that the semantically-similar patches ought to share their features, 

% It is observed from the Fig. \ref{vis} (a) that even the patches are similar, the vision token labels are still different. Fig. \ref{vis} (b) shows the corrupted image with 75\% random mask. The inter-patch similarity is shown in the form of heat map in the Fig. \ref{vis} (c). Take the patch located in the edge of the car as an example, the heat map of patch similarity obviously show the edges have brighter color, which demonstrate the model can predict and capture the semantics even based on the corrupted image. We further display the top-5 semantically-similar patch with respect to the selected masked patch. The eased and refined multi-choice label provides more  
% diverse answers in a probabilistic manner.

\section{Conclusion}

In this paper, we propose the \emph{mc}-BEiT, \textit{i.e.,} multi-choice discretization for improving image BERT pre-training. Instead of adopting the unique label signals from the tokenizer, we introduce an eased and refined objective for providing multi-choice answers. Extensive experiments are conducted to evaluate the performances of our method. The empirical results show that \emph{mc}-BEiT achieves the state-of-the-art performances on various tasks, such as image classification, semantic/instance segmentation, and objection detection.

\section*{Acknowledgement}
\label{acknowlegement}
This work was supported by the National Natural Science Foundation of China under Grant 62088102, and in part by the PKU-NTU Joint Research Institute (JRI) sponsored by a donation from the Ng Teng Fong Charitable Foundation.

% . Besides predicting the hard label as in BEiT, we introduce the inter-patch regularization into the reconstructed objective. Specially, for each masked patch, the relationship among patches is considered to guide the masked tokens to be reconstructed as well as learn more global structure. 
% Extensive experiments on various benchmarks demonstrate that our method achieves state-of-the-art performances compared to other methods, such as ImageNet-1K, 
% COCO, and ADE20K benchmarks.

\clearpage
% ---- Bibliography ----
%
% BibTeX users should specify bibliography style 'splncs04'.
% References will then be sorted and formatted in the correct style.
%
\bibliographystyle{splncs04}
\bibliography{eccv2022submission}

\newpage
\appendix
\section{Implementation Details}
In the appendix, we provide the specific hyper-parameters of the experiments in our paper, including pre-training on ImageNet-1K and fine-tuning on different downstream tasks.
\subsection{Configuration for pre-training}

The vision Transformers are pre-trained on the large-scale dataset ImageNet-1K \cite{imagenet1k} and the configurations are summarized in Tab. \ref{pre-training}. The implementation of the vision Transformers, \textit{i.e.,} ViT-Base/16 and ViT-Large/16, follows \cite{deit} for fair comparisons and the training recipe is based on BEiT \cite{beit}.

\begin{table}[h]
\small
\centering
\caption{Configurations for pre-training.}
\label{pre-training}
%\resizebox{\textwidth}{!}{
\setlength{\tabcolsep}{5mm}{
\begin{tabular}{l|cc}
\toprule[1pt]
Configuration        & ViT-Base/16            & ViT-Large/16           \\ \hline
Layers                 & 12                     & 24                     \\
Hidden size            & 768                    & 1024                   \\
FFN inner hidden size  & 3072                   & 4096                   \\
Attention heads        & 12                     & 16                     \\
Attention head size    & \multicolumn{2}{c}{64}                          \\
Patch size             & \multicolumn{2}{c}{$16\times 16$}   \\
\hline
Training epochs        & \multicolumn{2}{c}{800}                         \\
Batch size             & \multicolumn{2}{c}{2048}                        \\
Adam $\epsilon$               & \multicolumn{2}{c}{1e-8}                        \\
Adam $\beta$                   & \multicolumn{2}{c}{(0.9, 0.98)}                 \\
Peak learning rate     & \multicolumn{2}{c}{1.5e-3}                     \\
Minimal learning rate  & \multicolumn{2}{c}{1e-5}                        \\
Learning rate schedule & \multicolumn{2}{c}{Cosine}                      \\
Warmup epochs          & \multicolumn{2}{c}{10}                          \\
\hline
Gradient clipping      & 3.0                    & 1.0                    \\
%Dropout                & \multicolumn{2}{c}{\XSolid}      \\
Dropout                & \multicolumn{2}{c}{None}      \\
Stoch. depth           & \multicolumn{2}{c}{0.1}                         \\
Weight decay           & \multicolumn{2}{c}{0.05}                        \\
\hline
Data Augment           & \multicolumn{2}{c}{RandomResizeAndCrop}         \\
Input resolution       & \multicolumn{2}{c}{$224\times224$} \\
%Color jitter           & \multicolumn{2}{c}{0.4}                        
\bottomrule[1pt]
\end{tabular}
}
\end{table}

\subsection{Configuration for fine-tuning}

\subsubsection{Classification task on ImageNet-1K} 

For the classification task, the fully-connected layer is employed as the classifier after the average pooling of the feature embeddings. The fine-tuning configurations on ImageNet-1K for different backbone architectures are listed in Tab. \ref{finetuning}.
\begin{table}[h]
\small
% \vspace{3pt}
\centering
\caption{Configurations for fine-tuning on ImageNet-1K.}
\label{finetuning}
%\resizebox{\textwidth}{!}{
\setlength{\tabcolsep}{5mm}{
\begin{tabular}{l|cc}
\toprule[1pt]
Configuration                & ViT-Base/16            & ViT-Large/16           \\ \hline
Peak learning rate             & \multicolumn{2}{c}{\{2e-3,3e-3,4e-3,5e-3\}}     \\
Fine-tuning epochs             & 100                    & 50                     \\
Batch size                     & \multicolumn{2}{c}{1024}                                 \\
Warmup epochs                  & 20                     & 5                      \\
Layer-wise learning rate decay & 0.65                   & 0.75                   \\
Adam $\epsilon$                          & \multicolumn{2}{c}{1e-8}                        \\
Adam $\beta$                        & \multicolumn{2}{c}{(0.9, 0.999)}                \\
Minimal learning rate          & \multicolumn{2}{c}{1e-6}                        \\
Learning rate schedule         & \multicolumn{2}{c}{Cosine}                      \\ \hline
Repeated Aug                   & \multicolumn{2}{c}{None}                        \\
Weight decay                   & \multicolumn{2}{c}{0.05}                        \\
Label smoothing                & \multicolumn{2}{c}{0.1}                         \\
Stoch. depth                   & \multicolumn{2}{c}{0.1}                         \\
Dropout                        & \multicolumn{2}{c}{None}                        \\
Gradient clipping              & \multicolumn{2}{c}{None}                        \\ \hline
Erasing prob.                  & \multicolumn{2}{c}{0.25}                        \\
Input resolution               & \multicolumn{2}{c}{$224\times224$} \\
Rand Augment                   & \multicolumn{2}{c}{9/0.5}                       \\
Mixup prob.                    & \multicolumn{2}{c}{0.8}                         \\
Cutmix prob.                   & \multicolumn{2}{c}{1.0}            \\             
Color jitter           & \multicolumn{2}{c}{0.4}        \\               
\bottomrule[1pt]
\end{tabular}
% \vspace{3pt}
}
\end{table}

\subsubsection{Object detection and instance segmentation} 

We adopt the implementation of \cite{benchmarking,mimdet} to verify our performances of object detection and instance segmentation on COCO. Tab. \ref{tunecoco} summarizes the configurations for fine-tuning on COCO. The training recipe of models with intermediate fine-tuning is the same as the pre-training only version. ViT-B \cite{vit} is adopted as the backbone and Mask-RCNN \cite{maskrcnn} is used as the task head.

Besides, we also provide another experiment result following the implementation in iBOT \cite{ibot} in Tab. \ref{coco}. Because these experiments are not conducted on BEiT, we conduct the experiments following iBOT \cite{ibot} and the results of BEiT \cite{beit} are based on our re-implementation. In order to adapt to the multi-scale strategy, we use absolute
position embedding and interpolate it for different image resolutions. ViT-B \cite{vit} is adopted as the backbone and Cascaded Mask-RCNN \cite{maskrcnn,cascade} is used as the task head.

\begin{table}[H]
% \small
\centering
%\resizebox{\textwidth}{!}{
\caption{Configurations for fine-tuning on COCO.}
\setlength{\tabcolsep}{5mm}{
\begin{tabular}{l|c}
\toprule[0.7pt]
Configuration & ViT-Base/16  \\
\hline
Fine-tuning epochs             & \multicolumn{1}{c}{25}          \\
Peaking learning rate          & \multicolumn{1}{c}{8e-5}         \\
Learning rate decay            & \multicolumn{1}{c}{cosine}  \\
Adam $\epsilon$                & \multicolumn{1}{c}{1e-8}         \\
Adam $\beta$                   & \multicolumn{1}{c}{(0.9, 0.999)} \\
\hline
Dropout                        & \multicolumn{1}{c}{None} \\
Stoch. depth                   & \multicolumn{1}{c}{0.1} \\
Weight decay                   & \multicolumn{1}{c}{0.1} \\
Batch size                     & \multicolumn{1}{c}{64} \\
\hline
Input size                     & \multicolumn{1}{c}{$1024\times1024$} \\
Position embedding             & \multicolumn{1}{c}{Abs. + Rel.}     \\
Augmentation                   & \multicolumn{1}{c}{LSJ(0.1, 2.0)} \\
\bottomrule[0.7pt]
\end{tabular}
}
% \vspace{3pt}
\label{tunecoco}
\end{table}

\begin{table}[h]
% \setlength{\abovecaptionskip}{0.cm}
% \small
\centering
\caption{We provide another experiment results of object detection and instance segmentation on COCO following the implementation of iBOT\cite{ibot}. Intermediate fine-tuning denotes the model is further fine-tuned on ImageNet-1K. Cascaded Mask R-CNN and 1$\times$ training schedule are adopted. }
\label{coco}
%\resizebox{\textwidth}{!}{
\setlength{\tabcolsep}{3.5mm}{
\begin{tabular}{l|ccc}
%\hline
\toprule[1pt]
\multicolumn{1}{l|}{\multirow{2}{*}{Method}} &
  \multicolumn{1}{c}{\multirow{2}{*}{Reference}} &
  \multicolumn{1}{c}{Object Det.} &
  Instance Seg. \\ %\cline{3-4} 
\multicolumn{1}{l|}{} &
  \multicolumn{1}{c}{} &
  \multicolumn{1}{c}{$\text{AP}^{b}$} &
  \multicolumn{1}{c}{$\text{AP}^{m}$} \\ \hline
% \multicolumn{1}{l|}{Method} & Reference & $\text{AP}^{b}$ & $\text{AP}^{m}$ \\ \hline
Supervised \cite{deit} & ICML 2021 & 47.9 & 42.9\\
MoCo v3 \cite{mocov3} & CVPR 2021 & 47.9 & 42.7 \\
DINO \cite{dino} & ICCV 2021  & 50.1 & 43.4 \\
iBOT \cite{ibot} & ICLR 2022 & 51.2 & 44.2 \\
BEiT \cite{beit} & ICLR 2022 & 49.6 & 42.8 \\
Ours & this paper & 50.1 & 43.1 \\ \hline
+Intermediate Fine-tuning \\ \hline
BEiT \cite{beit} & ICLR 2022 & 50.7 & 43.8 \\
Ours & this paper & \textbf{51.2} & \textbf{44.3} \\

%\hline
\bottomrule[1pt]
\end{tabular}
}
%}
\end{table}

\subsubsection{Semantic segmentation on ADE20K:} 
For the semantic segmentation experiments on ADE20K \cite{ade20k}, we follow the implementation of BEiT \cite{beit} and adopt UperNet  \cite{upernet} as the task layer. ViT-B \cite{vit} is adopted as the default backbone and UPerNet \cite{upernet} is used as the task head. Tab. \ref{tuneade20k} summarizes the configurations for fine-tuning on ADE20k. Because the pre-training process does not introduce the instance discrimination, the performance can be further improved after intermediate fine-tuning on ImageNet-1K according to BEiT \cite{beit}. we also evaluate the performances after intermediate fine-tuning, where the pre-trained models have been fine-tuned on ImageNet-1K. For the models with intermediate fine-tuning, the training recipe is the same as the pre-training only version.

%We conduct the downstream experiments on ADE20K and COCO, the configurations are summarized in Tab. \ref{tuneade20k} and Tab. \ref{tunecoco}, respectively. 

\begin{table}[t]
\small
\centering
\caption{Configurations for fine-tuning on ADE20k.}
\label{tuneade20k}
%\resizebox{\textwidth}{!}{
\setlength{\tabcolsep}{5mm}{
\begin{tabular}{l|cc}
\toprule[1pt]
Configuration                & ViT-Base/16  \\ \hline
Peaking learning rate          & 8e-5        \\
Fine-tuning steps              & 160000         \\
Batch size                     & 16           \\
Adam $\epsilon$                          & 1e-8         \\
Adam    $\beta$                       & (0.9, 0.999) \\
Layer-wise learning rate decay & 0.9         \\
Minimal learning rate          & 0            \\
Learning rate schedule         & Linear       \\
Warmup steps                   & 1500         \\ \hline
Dropout                        & None         \\
Stoch. depth                   & 0.1          \\
Weight decay                   & 0.05         \\ \hline
Input resolution               & 512$\times$512      \\
Position embedding             & Relative     \\
Position embedding interpolate & Bilinear    \\        
%Color jitter           & \multicolumn{2}{c}{0.4}                        
\bottomrule[1pt]
\end{tabular}
}
\end{table}

\end{document}